\documentclass{article}

\PassOptionsToPackage{table}{xcolor}
\usepackage{iclr2027_conference,times}

\usepackage[utf8]{inputenc}
\usepackage[T1]{fontenc}
\usepackage{booktabs}
\usepackage{amsfonts}
\usepackage{nicefrac}
\usepackage{microtype}
\usepackage{multirow}
\usepackage{graphicx}
\usepackage{adjustbox}
\usepackage{capt-of}
\usepackage{amsmath}
\usepackage{bm}
\usepackage{tabularx}
\usepackage[abs]{overpic}
\usepackage{url}
\usepackage{hyperref}
\hypersetup{
  hidelinks,
  pdfauthor={Haoran Qin, Renlong Wu, Tianyu Huang, Yukang Ding, Hui Li, Wangmeng Zuo},
  pdftitle={CompAdapt: Adaptable Composite Motion Modeling for Physics-Consistent Text-to-Video Generation}
}

\RequirePackage{xspace}
\makeatletter
\DeclareRobustCommand\onedot{\futurelet\@let@token\@onedot}
\def\@onedot{\ifx\@let@token.\else.\null\fi\xspace}

\def\eg{\emph{e.g}\onedot} 
\def\ie{\emph{i.e}\onedot} 
 
\def\etc{\emph{etc}\onedot}

\makeatother

\definecolor{Blue}{RGB}{0,176,240}
\definecolor{Green}{RGB}{0,176,80}

\usepackage[capitalize]{cleveref}
\crefname{section}{Sec.}{Secs.}
\Crefname{section}{Section}{Sections}
\Crefname{table}{Table}{Tables}
\crefname{table}{Tab.}{Tabs.}
\Crefname{equation}{Equation}{Equations}
\crefname{equation}{Eqn.}{Eqns.}

\newcommand{\best}[1]{\textbf{#1}}
\newcommand{\secondbest}[1]{\underline{#1}}

\title{CompAdapt: Adaptable Composite Motion Modeling for Physics-Consistent Text-to-Video Generation}

\author{%
  Haoran Qin$^{1}$, Renlong Wu$^{1}$, Tianyu Huang$^{1}$, Yukang Ding$^{2}$, Hui Li$^{1}$, Wangmeng Zuo$^{1}$ \\
  $^{1}$Harbin Institute of Technology, China \\
  $^{2}$Taobao, Alibaba Group, China \\
}

\iclrfinalcopy

\begin{document}

\maketitle
\lhead{Preprint}

\begin{abstract}
While diffusion-based text-to-video (T2V) models have demonstrated impressive capability in generating realistic and temporally coherent videos, they often fail to respect fundamental physical dynamics. 
Although recent physics-constrained methods incorporate explicit dynamics priors to improve physical plausibility, they remain limited to simple single-type motions, depend on manually specified parameters, and struggle to generalize to unseen physical laws.
%
In this work, we propose CompAdapt, a physics-consistent T2V framework for adaptable generation across complex real-world scenarios. 
It extends neural dynamics modeling beyond single-type motions to encompass composite physical behaviors, including coupled motions, multi-stage transitions, and multi-object collisions.
%
Furthermore, CompAdapt translates natural language prompts into structured physical semantics, enabling end-to-end specification of motion types, temporal relations, and initial physical parameters. 
To generalize to novel physical environments, CompAdapt introduces dynamics-aware prior matching, achieving one-shot adaptation without retraining the core dynamics module.
In addition, a physics-aware latent feature fusion module improves visual fidelity under fast and complex motion. 
Experiments on physics-focused T2V benchmarks demonstrate that CompAdapt improves physical consistency over both general T2V models and physics-constrained baselines, while preserving high visual quality and adaptability to unseen dynamics. The project page is available at \url{https://makapic.github.io/CompAdapt/}.
\end{abstract}

\section{Introduction}
Diffusion-based text-to-video (T2V) models~\citep{ho2022videodiffusion,ho2022imagenvideo,singer2023makeavideo,yang2025cogvideox,bartal2024lumiere,chen2023videocrafter1,chen2024videocrafter2,guo2023animatediff} have recently achieved impressive progress in generating realistic and temporally coherent videos, supporting applications in content creation, simulation, and immersive media.
However, visual realism alone does not guarantee physical correctness. 
Most T2V models are trained to reproduce motion patterns from large-scale video data, where motion is learned mainly through visual correlations rather than explicit physical dynamics~\citep{guo2025t2vphysbench,chen2025physicalcoherencebenchmark}. 
As a result, they may generate videos that appear plausible at first glance but violate basic physical rules, such as gravity, force-induced acceleration, collision response, and coherent transitions between different motion components.

\begin{figure}[!t]
  \centering
  \includegraphics[width=\textwidth]{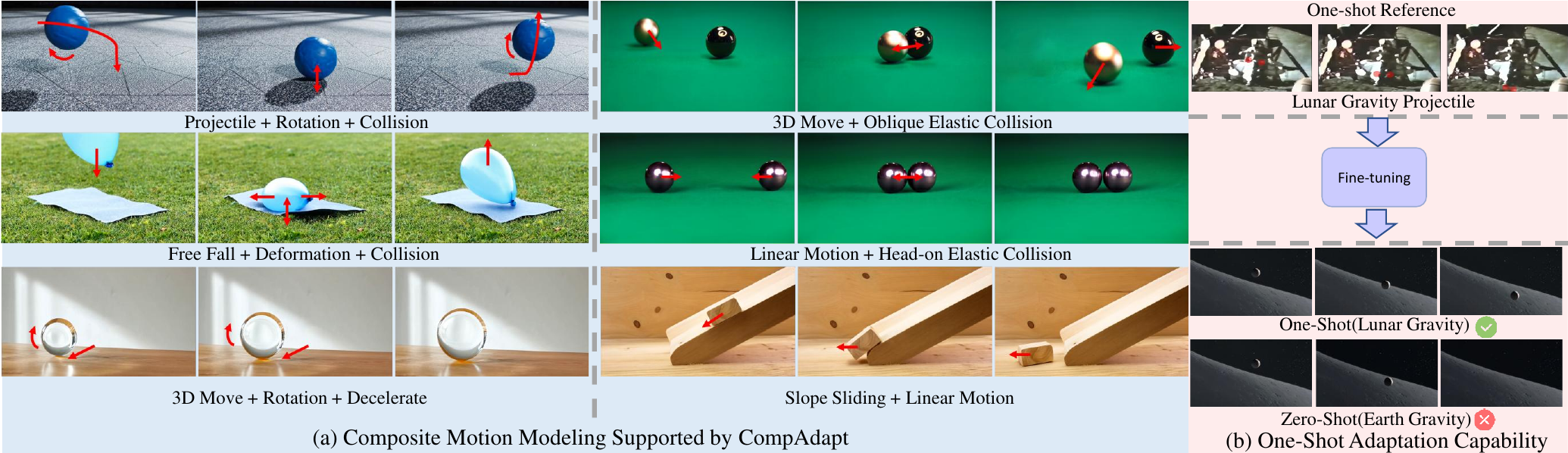}
  \caption{CompAdapt generates physically-consistent composite motion videos from texts, with (a) unified composite motion modeling and (b) efficient one-shot adaptation to novel physical laws.}
  \label{fig:compadapt-teaser}
  \vspace{-6mm}
\end{figure}

To address physical inconsistency in T2V generation, recent works incorporate explicit physical priors to guide the generation process with physical dynamics constraints.
NewtonGen~\citep{yuan2025newtongen} is a representative one, which builds upon physics-informed neural ordinary differential equations (ODEs)~\citep{chen2018neuralode,raissi2019pinn} and introduces a Neural Newtonian Dynamics (NND) module to model basic Newtonian motion. 
This design improves the physical plausibility and enables controllable single-object motion synthesis. 
However, existing physics-constrained T2V methods are still limited in complex real-world scenarios.
First, they mainly focus on single-type fundamental motions and lack support for composite motion, where multiple motion components may occur in parallel, sequentially, or be interrupted by discrete dynamic events such as collisions~\citep{gillman2025forceprompting,gillman2026goalforce,zhang2026physrvg}. 
Second, their motion parameters are typically manually specified, which limits their practicality in open-world generation. 
Third, these methods have limited adaptability to novel physical laws.
When encountering out-of-distribution environments, such as lunar gravity or low-friction surfaces, they usually require retraining the core dynamics module with large-scale physics-clean synthetic data, lacking efficiency for adaptation~\citep{chen2018neuralode,rubanova2019latentode,greydanus2019hnn,cranmer2020lnn}.
Therefore, physics-aware T2V generation requires a more flexible framework for handling complex real-world scenarios, which should flexibly compose multiple motion dynamics and rapidly adapt to unseen physical conditions.
Based on these observations, we propose CompAdapt, an adaptable physics-consistent T2V generation framework for complex real-world scenarios, as shown in Figure~\ref{fig:compadapt-teaser}. 
CompAdapt provides three key capabilities. 
First, it represents complex real-world motion through parallel and sequential composition, supporting coupled motions, multi-stage transitions, and multi-object collisions beyond single-type fundamental motion. 
Second, it translates natural language prompts into structured physical semantics, including motion categories, temporal relations, and initial physical parameters. 
Third, it enables one-shot adaptation to unseen physical laws through dynamics-aware prior matching, allowing efficient adaptation to conditions such as lunar gravity or low-friction surfaces.

Specifically, CompAdapt realizes these capabilities through a language-to-dynamics-to-video pipeline. 
A text-to-physical parser first converts the input prompt into a standardized motion sequence and initial physical state. 
Based on an augmented 10-dimensional physical state representation, neural ODE-based dynamics modules predict object-level motion trajectories. 
The Motion Type Composite Module (MTC) then composes parallel and multi-stage motions into a coherent physical trajectory, while the Multi-Object Interaction Module (MOT) handles discrete multi-object interactions such as collisions.
The predicted trajectory is further used to guide a physics-aware video generator with state-dependent latent feature propagation. 
For novel physical environments, dynamics-aware prior matching selects and adapts the most compatible dynamics module from a single observation.

We evaluate CompAdapt on physics-focused benchmarks covering motion parsing, composite motion generation, multi-object collision, and adaptation to unseen physical laws. 
Extensive experiments show that CompAdapt consistently improves physical consistency over both general T2V models and physics-constrained baselines. 
In particular, it better handles composite motions and collision dynamics, and can adapt to novel physical environments from one sample. 
These results demonstrate the effectiveness of CompAdapt for physics-consistent T2V generation in complex real-world scenarios.

The main contributions are summarized as follows:
\begin{itemize}
\item We propose CompAdapt, a physics-consistent T2V generation framework, which integrates prompt-based physical parameter parsing, composite dynamics modeling, and one-shot adaptation to novel physical laws for producing videos in complex physical scenarios.

\item We introduce a text-to-physical parsing and composite motion modeling scheme. It converts prompts into structured physical semantics and composes parallel, sequential, and collision motions into coherent physical trajectories.

\item We design a dynamics-aware one-shot adaptation strategy for unseen physical laws. By matching a single observation to the most compatible pre-trained dynamics module, CompAdapt can handle novel physical conditions with limited updates.

\item We conduct extensive experiments on physics-focused T2V generation benchmarks. The results demonstrate that CompAdapt achieves stronger physical consistency than general T2V models and physics-constrained baselines, while maintaining high visual fidelity.
\end{itemize}

\section{Related Work}

\subsection{Physics-Aware and Motion-controllable Video Generation}
Recent text-to-video models can generate visually impressive videos, but they often violate basic physical laws~\citep{huang2023vbench,huang2024vbenchpp}. 
Existing physics-aware generation methods improve physical consistency through simulation-based pipelines, embedded dynamics priors, or post-hoc optimization~\citep{zhang2024physdreamer,liu2024physgen,xie2024physgaussian,zhang2026physrvg,guo2025t2vphysbench,chen2025physicalcoherencebenchmark}. 
NewtonGen~\citep{yuan2025newtongen} is closely related to our work, as it introduces Neural Newtonian Dynamics for controllable physical motion generation. 
However, it mainly focuses on fundamental motion types and relies on manually specified physical parameters, limiting its applicability to composite motions, collisions, and unseen physical environments.
Motion-controllable video generation provides another important direction for improving generation controllability. 
Representative methods guide video generation with trajectories, drag signals, camera motion, or motion prompts~\citep{guo2023animatediff,yin2023dragnuwa,geng2025motionprompting,shi2024motioni2v,li2025magicmotion,liao2025motionagent,ling2025motionclone,he2024cameractrl,he2024mojito}. 
These methods improve motion alignment but generally do not explicitly model physical dynamics. 
We build upon Wan-Move~\citep{chu2025wanmove} and introduce physics-aware feature propagation guided by predicted physical states.

\subsection{Neural Dynamics Modeling and Physical Evaluation}
Neural dynamics models, including PINNs, Neural ODEs, Hamiltonian Neural Networks, and Lagrangian Neural Networks, have been widely studied for modeling physical systems~\citep{raissi2019pinn,chen2018neuralode,rubanova2019latentode,greydanus2019hnn,cranmer2020lnn}. 
Relational and graph-based dynamics models further extend this idea to multi-object interaction modeling~\citep{kipf2018nri,sanchezgonzalez2020graphsims}. 
Despite their progress, applying neural dynamics to open-ended T2V generation remains difficult due to the need for text-conditioned parameter parsing, composite motion composition, and adaptation to novel physical laws.
We address these limitations with text-to-physical parsing, composite dynamics modeling, and one-shot adaptation.

\section{Methodology}
\label{sec:method}
\begin{figure*}[t]
    \centering
    \includegraphics[width=\textwidth]{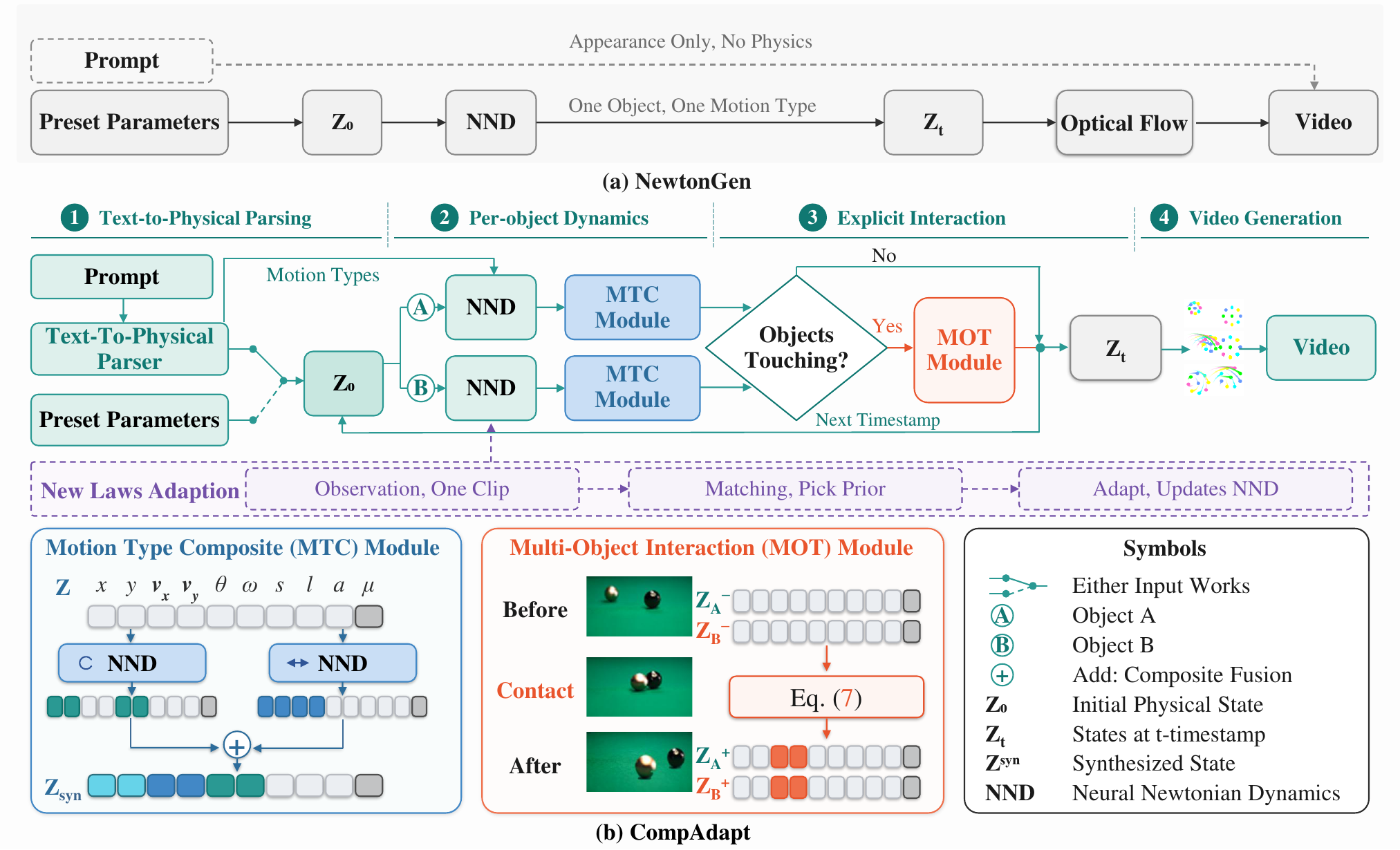}
    \caption{Overview of CompAdapt. (a) NewtonGen uses one NND with preset parameters for a single object and motion type. (b) CompAdapt adds text-to-physical parsing, composite motion modeling (MTC), multi-object interaction (MOT), and one-shot adaptation. The resulting state $Z(t)$ guides video generation.}
    \label{fig:compadapt_pipeline}
    \vspace{-4mm}
\end{figure*}

Given a text prompt that describes a physical scene and its intended motion, our goal is to generate a visually realistic and physically consistent video, which presents four challenges.
First, a free-form prompt does not directly provide the physical quantities required by a dynamics model, such as velocity, acceleration, mass, or collision timing. 
Second, real-world motion often contains multiple components that occur in parallel, unfold sequentially, or interact through discrete events such as collisions. 
Third, trajectory-guided video generators tend to produce texture distortion under complex motion. 
This occurs because naively replicating first-frame features along a motion path ignores how deformation and rotation modulate visual appearance across frames. 
Fourth, pre-trained dynamics modules learned under standard physical conditions no longer match when the environment changes, such as under lunar gravity or on low-friction surfaces, calling for adaptation to novel physical laws.
To address these challenges, we propose CompAdapt, as shown in~\cref{fig:compadapt_pipeline}. 
CompAdapt first converts the input text into structured physical semantics, then models composite physical motion into trajectories, and finally renders a physics-aware video guided by the predicted trajectory.
In addition, we introduce a one-shot adaptation paradigm for novel physical laws. 
We first formulate the physical state representation in~\cref{sec:phys_state} and then detail the text-to-physical parser in Sec.~\cref{sec:text2phys}. 
Next, we describe composite motion modeling in~\cref{sec:comp_motion}. 
Finally, we present the physics-aware video generation pipeline in~\cref{sec:video_gen} and the one-shot adaptation paradigm in~\cref{sec:oneshot}.

\subsection{Physical State Representation}
\label{sec:phys_state}
NewtonGen~\citep{yuan2025newtongen} introduces a physics-informed neural ODE framework for modeling single-object continuous motion, where the physical state is represented as a 9-dimensional latent vector, \ie,
\[
\mathbf{Z} = [x, y, v_x, v_y, \theta, \omega, s, l, a].
\]
$(x, y)$ and $(v_x, v_y)$ are the 2D centroid position and velocity, $\theta$ and $\omega$ the angle and angular velocity, and $s$, $l$, and $a$ the shortest side, longest side, and projected area.
Because momentum transfer during multi-object collisions depends on mass, a quantity not determined by kinematics or geometry alone, we augment the physical state representation with an explicit mass dimension $\mu$, \ie,
\[
\mathbf{Z}' = \{\mathbf{Z}, \mu\}.
\]
For notational simplicity, we use $\mathbf{Z}$ to denote $\mathbf{Z}'$ hereafter.
Following NewtonGen~\citep{yuan2025newtongen}, the temporal dynamics of each element $z$ in $\mathbf{Z}$ follows a second-order ODE that combines linear physics terms with a nonlinear residual, \ie, 
\begin{equation}
a_z \ddot{z} + b_z \dot{z} + c_z z + d_z + \mathrm{MLP}(\mathbf{Z}) = 0,
\label{eq:ode}
\end{equation}
where $a_z, b_z, c_z, d_z$ are learnable parameters capturing the dominant Newtonian dynamics, and an MLP models the unknown residuals.
Given an initial state $\mathbf{Z}_0$, the future state $\mathbf{Z}_t$ is obtained via a neural ODE integrator~\citep{chen2018neuralode}, \ie,
\begin{equation}
\mathbf{Z}_t = \mathbf{Z}_0 + \int_{t_0}^t \mathrm{Func}(\mathbf{Z}(\tau))\, d\tau.
\label{eq:nnd}
\end{equation}
$\mathrm{Func}(\mathbf{Z}(\tau))$ denotes the set of individual $\frac{\mathrm{d}z}{\mathrm{d}t}$ ODEs governing each element, and $\mathbf{Z}_0 = \mathbf{Z}(t_0)$ is the known initial physical state at time $t_0$.

\subsection{Text-to-Physical Parser}
\label{sec:text2phys}
Real-world composite motion is inherently hierarchical, \ie, motions at a timestamp are executed in parallel, while consecutive timestamps unfold sequentially.
Translating such motion from a free-form text prompt into a structured physical state further requires two distinct capabilities, \ie, recognizing motion semantics (\eg, category and temporal order) and estimating numerical physical parameters (\eg, velocity and mass). 
We design a dual-LLM parser based on Qwen2.5-7B-Instruct~\citep{yang2024qwen25}, \ie, $\text{LLM}_{\text{seq}}$ and $\text{LLM}_{\text{param}}$, to decouple these objectives.
Let $\text{P}$ denote the input text prompt.
$\text{LLM}_{\text{seq}}$ extracts the motion components and determines their temporal relations, where each motion is classified into one of 12 predefined categories, \eg, uniform motion, accelerated motion, and collision.
It outputs a set of motion sequences $\mathcal{S}$ for $N$ objects, \ie,
\begin{equation}
\mathcal{S} = \text{LLM}_{\text{seq}}(\text{P}) = \{\mathbf{S}^{(1)}, \mathbf{S}^{(2)}, \ldots, \mathbf{S}^{(N)}\},
\label{eq:LLMseq}
\end{equation}
where $\mathbf{S}^{(n)} = (\mathcal{M}^{(n)}_1, \ldots, \mathcal{M}^{(n)}_T)$ is the motion sequence for the $n$-th object, and $T$ is the number of timestamps.
$\mathcal{M}^{(n)}_i$ is the synchronous motion components in the $i$-th timestamp, which can be written as,
\begin{equation}
\mathcal{M}^{(n)}_i = \{m^{(n)}_{i,1}, \ldots, m^{(n)}_{i,o_i}\}.
\end{equation}
$m^{(n)}_{i,j}$ is the $j$-th component and $o_i$ is the number of components at timestamp $i$.
$\text{LLM}_{\text{param}}$ takes $\text{P}$ and $\mathcal{S}$ as inputs and estimates the physical state $\mathcal{Z}_0$ for the initial timestamp, which can be written as,    
\begin{equation}
\mathcal{Z}_0 = \text{LLM}_{\text{param}}(\text{P}, \mathcal{S}) = \{\mathbf{Z}^{(1)}_0, \mathbf{Z}^{(2)}_0, \ldots, \mathbf{Z}^{(N)}_0\}.
\label{LLMparam}
\end{equation}
$\mathbf{Z}^{(n)}_0$ is the physical state for the $n$-th object.

\subsection{Composite Motion Modeling}
\label{sec:comp_motion}
Given per-object motion sequences $\mathbf{S}^{(n)}$ and their initial physical states $\mathbf{Z}^{(n)}_0$, we instantiate the corresponding executable neural dynamics for each object.
Specifically, each motion component $m^{(n)}_{i,j} \in \mathcal{M}^{(n)}_i$ activates its category-specific pre-trained NND module $\mathcal{D}_{\mathrm{cat}(m^{(n)}_{i,j})}$, which takes $\mathbf{Z}^{(n)}_0$ as the initial condition and propagates it into a time-continuous physical state $\mathbf{Z}^{(n)}_{i,j}(t)$.
To compose $\mathbf{Z}^{(n)}_{i,j}(t)$ into physically consistent object-level trajectories, we introduce a unified compositional framework in which the Motion Type Composite Module (MTC) fuses concurrent motion components at each timestamp, and the Multi-Object Interaction Module (MOT) resolves discrete multi-object contacts when interactions occur.

\textbf{Motion Type Composite Module.}
When multiple motion components act on an object simultaneously, their corresponding NND modules operate on mostly disjoint subsets of $\mathbf{Z}$-space variables, so simultaneous motions such as rotation paired with deceleration naturally superpose at the kinematic level within the physical state representation.
Furthermore, since all components share the same initial state $\mathbf{Z}_{\mathrm{init}}$, this composition rule is strictly commutative.
This additivity and commutativity motivate an additive composition scheme for parallel motion components.
Concretely, at timestamp $t_i$ the object is subject to $o_i$ parallel components $\mathcal{M}^{(n)}_i$.
Summing these rollouts would count the force-free motion once per component (\eg the $v_0t$ drift is duplicated under concurrent accelerations). We therefore count it once and add each component's particular response. Let $F_{\mathrm{free}}$ be the force-free field, which keeps the linear and angular velocities constant and the shape fixed (Newton's first law on every block), with rollout $\mathbf{Z}_{\mathrm{free}}(t)=\Phi_t^{F_{\mathrm{free}}}(\mathbf{Z}_{\mathrm{init}})$. The particular response $\mathbf{Z}^{(n)}_{i,j}(t)-\mathbf{Z}_{\mathrm{free}}(t)$ then vanishes at $t=0$.
The composed state is then
\begin{equation}
\mathbf{Z}^{\mathrm{syn}}(t)
=
\mathbf{Z}_{\mathrm{free}}(t)
+
\sum_{j=1}^{o_i}
\mathbf{r} \odot \bigl(\mathbf{Z}^{(n)}_{i,j}(t) - \mathbf{Z}_{\mathrm{free}}(t)\bigr).
\label{eq:syn_unit}
\end{equation}
Here $\odot$ is element-wise multiplication, and $\mathbf{r} = [1,1,1,1,1,1,1,1,1,0]$ zeros the mass coordinate, which does not change under composition.
On disjoint blocks Eq.~\eqref{eq:syn_unit} reduces to the displacement sum. On shared coordinates with affine-linear dynamics it gives the exact free-plus-particular superposition.
Under the temporal chaining paradigm, $\mathbf{Z}_{\mathrm{init}}$ is set to $\mathbf{Z}^{(n)}_0$ at the first timestamp, and to the synthesized terminal state $\mathbf{Z}^{\mathrm{syn}}(t_{i-1})$ from the preceding timestamp for all subsequent steps. This ensures continuous trajectory concatenation across the full sequence.

\textbf{Multi-Object Interaction Module.}
While the MTC handles smooth continuous motion, collisions introduce instantaneous, discontinuous state transitions that fall outside the scope of standard continuous NND evolution.
To capture these discrete events, we introduce a collision dynamics module $\mathcal{D}_{\mathrm{col}}$, a separately trained NND that maps pre-collision physical states to post-collision physical states, which can be written as,
\begin{equation}
(\mathbf{Z}_{t_c}^{(A,+)},\; \mathbf{Z}_{t_c}^{(B,+)}) = \mathcal{D}_{\mathrm{col}}(\mathbf{Z}_{t_c}^{(A,-)},\; \mathbf{Z}_{t_c}^{(B,-)}).
\label{eq:collision}
\end{equation}
$\mathbf{Z}_{t_c}^{(A,-)}$ and $\mathbf{Z}_{t_c}^{(B,-)}$ denote the pre-collision physical states of objects $A$ and $B$ at collision time $t_c$. $\mathbf{Z}_{t_c}^{(A,+)}$ and $\mathbf{Z}_{t_c}^{(B,+)}$ are the post-collision states.
The contact time $t_c$ is the first timestamp at which the minimum distance between the two convex object shapes falls below a threshold $\epsilon$ (Eq.~\eqref{eq:boundary_dist}); otherwise each object continues to evolve independently.
After a collision, the two objects continue from $\mathbf{Z}_{t_c}^{(A,+)}$ and $\mathbf{Z}_{t_c}^{(B,+)}$ until the next collision or the end of the motion sequence.
Formal derivations of both rules, with exactness conditions, invariant preservation, and a coupling error bound, are given in \cref{sec:app-comp-proof}.

\subsection{Physics-Aware Video Generation}
\label{sec:video_gen}

The composite motion modeling module outputs a full sequence of Z-space physical states with guaranteed physical consistency. As the rendering branch of our decoupled framework, the video generation module translates these abstract kinematic states into realistic video frames, allowing independent optimization of physical accuracy and visual quality.

We build our rendering pipeline on Wan-Move~\citep{chu2025wanmove}, a trajectory-conditioned video diffusion model, as per-frame centroid trajectories can be directly projected from physical states. Its core motion guidance mechanism relies on latent feature replication. Given the first-frame latent feature map \(\mathbf{F}_0\) and per-frame point coordinates \((u_f, v_f)\) projected from each \(\mathbf{Z}_f\), it copies feature values from the initial trajectory position to all subsequent frames in a hard manner, formulated as,

\begin{equation}
\mathbf{F}_f(u_f, v_f, :) = \mathbf{F}_0(u_0, v_0, :).
\label{eq:hard_copy}
\end{equation}
Here \(\mathbf{F}_f\) denotes the latent condition feature map at frame \(f\). However, this global hard copying scheme only models translational displacement and leaves rotation- and scaling-induced distortion uncorrected.
To mitigate this issue, we inject physical geometric priors into the diffusion feature propagation process.
Specifically, we design two complementary modules to implement this idea. 
One is a spatial warp module dedicated to explicit geometric transformation modeling, and the other is an adaptive blending module for fine-grained artifact compensation.  

\textbf{Spatial Warp Module.} Instead of global feature copying, we adopt spatially transformed local patch propagation. For each decoder block, a local latent patch \(\mathbf{H}_0\) is extracted from \(\mathbf{F}_0\) centered at the initial object position \((u_0, v_0)\). For frame \(f\), this patch is warped according to the transformation matrix, \ie, 
\begin{equation}
\mathbf{M}_f = \mathbf{R}(\theta_f - \theta_0) \cdot \mathrm{diag}(\tfrac{s_f}{s_0},\, \tfrac{l_f}{l_0}),
\label{eq:spatial_transform}
\end{equation}
where \(\mathbf{M}_f\) denotes the spatial transformation matrix for frame \(f\), \(\mathbf{R}(\cdot)\) is the 2D rotation matrix, and \(\mathrm{diag}(\cdot)\) constructs a diagonal scaling matrix. This design explicitly models rotation and scaling during feature propagation.

\textbf{Adaptive Blending Module.} To further compensate for artifacts from latent-space scaling and out-of-plane appearance changes, we introduce a lightweight MLP that takes the relative dimensional change vector \(\boldsymbol{\rho}_f = (\tfrac{s_f}{s_0},\, \tfrac{l_f}{l_0},\, \tfrac{a_f}{a_0})\) as input and predicts a channel-wise gate \(\mathbf{G}\) to perform adaptive feature blending. The blending operation is formulated as,
\begin{equation}
\mathbf{h}_f = \mathbf{G} \odot \mathbf{M}_f(\mathbf{H}_0) + (\mathbf{1} - \mathbf{G}) \odot \tilde{\mathbf{h}}_f,
\label{eq:feature_blending}
\end{equation}
where \(\mathbf{h}_f\) is the final fused feature for frame \(f\), and \(\tilde{\mathbf{h}}_f\) is the native diffusion feature at the current denoising step. 
\subsection{One-Shot Adaptation to Novel Physical Laws}
\label{sec:oneshot}
One-shot adaptation starts from the observed motion. SAM2~\citep{ravi2024sam} tracks the object, and we read the state sequence $\{\hat{\mathbf{Z}}_f\}_{f=0}^{F}$ from its masks and trajectory.
Adapting a dynamics module to a novel physical environment must satisfy two core requirements.
First, selecting a strong initialization that matches the target dynamics to improve convergence efficiency and adaptation reliability.
Second, preserving the learned physical prior during fine-tuning to avoid overfitting to a single observation.
To address these requirements, we perform a two-step adaptation based on the extracted observation sequence with the Prior Matching Module and the Optimization Loss.

\textbf{Prior Matching Module.}
After that, we perform dynamics-aware prior matching to select the most suitable pre-trained dynamics module.
Let $\{\mathcal{D}_k\}_{k=1}^{K}$ denote the set of pre-trained single-object dynamics modules, where each module $\mathcal{D}_k$ is parameterized by $\phi_k^{0}$.
Starting from the observed initial state $\hat{\mathbf{Z}}_0$, each module predicts a trajectory, \ie,
\begin{equation}
\tilde{\mathbf{Z}}_{f}^{(k)} = \mathcal{D}_k(\hat{\mathbf{Z}}_0, f; \phi_k^{0}),
\quad f=0,\dots,F.
\end{equation}
We select the module $\mathcal{D}_{k^{*}}$ whose predicted trajectory best matches $\{\hat{\mathbf{Z}}_f\}$, which provides a better initialization for one-shot adaptation.
It can be written as,
\begin{equation}
k^{*} =
\arg\min_{k}
\frac{1}{F}
\sum_{f=0}^{F}
\left\|
\tilde{\mathbf{Z}}_{f}^{(k)}
-
\hat{\mathbf{Z}}_f
\right\|_2^2 .
\end{equation}

\textbf{Optimization Loss.}
Finally, we fine-tune $\mathcal{D}_{k^{*}}$ on the extracted trajectory with the proposed optimization loss, which can be written as,
\begin{equation}
\mathcal{L}_{\mathrm{adapt}}
=
\frac{1}{F}
\sum_{f=0}^{F}
\left\|
\mathcal{D}_{k^{*}}(\hat{\mathbf{Z}}_0, f; \phi)
-
\hat{\mathbf{Z}}_f
\right\|_2^2
+
\lambda
\left\|
\phi - \phi_{k^{*}}^{0}
\right\|_2^2 .
\label{eq:adapt_loss}
\end{equation}
$\phi_{k^{*}}^{0}$ denotes the original pre-trained parameters, and $\lambda$ controls the regularization strength.
The first term fits the observed trajectory, while the second term prevents the adapted model from drifting too far from the learned dynamics prior. 
CompAdapt employs the updated dynamics module to generate videos under the new physical condition.

\section{Experiments}
\label{sec:experiments}
\subsection{Experimental Settings}
\textbf{Datasets.} We construct four physics-focused datasets (\ie, \textit{Text2Phys}, \textit{CollidePhys}, \textit{CompoPhys}, and \textit{OODPhys}) from physical reasoning benchmarks (\ie, PHYRE~\citep{bakhtin2019phyre}, Physion~\citep{bear2021physion}, and I-PHYRE~\citep{li2023iphyre}) and large-scale T2V datasets (\ie, Panda-70M~\citep{chen2024panda70m} and  OpenVid-1M~\citep{nan2025openvid}), following the physics-clean simulation protocol in NewtonGen~\citep{yuan2025newtongen}.
\textit{Text2Phys} contains 10,000 text prompts with structured physical annotations, which are used to train the text-to-physical parser.
\textit{CollidePhys} has 1,000 two-object elastic-collision videos, 800 for training and 200 for testing the collision module.
\textit{CompoPhys} has 1,500 composite-motion videos, 1,200 for training and 300 for testing, with per-frame states for parallel, sequential, and hybrid compositions.
\textit{OODPhys} contains 34 controlled out-of-distribution scenarios and three real-world videos covering lunar free-fall, low-friction curling, and a damped pendulum.
The scenarios are organized into four difficulty tiers by their structural gap to the pre-trained dynamics modules: 22 near-domain (parameter shift), 5 mid-domain (partial structure shift), 5 far-domain (structural OOD), and 2 adversarial (capacity limit).
Scenario definitions and the sampling protocol are provided in \cref{sec:app-oodphys}.

\textbf{Implementation Details.}
We fine-tune a Qwen2.5-7B-Instruct~\citep{yang2024qwen25} with LoRA as the dual-branch text-to-physical parser.
The learning rate, LoRA rank and batch size are set to $2 \times 10^{-5}$, 8, and 8 respectively.
For the dynamics modules, we use the pre-trained models from NewtonGen~\citep{yuan2025newtongen}.
The collision-specific module is trained on \textit{CollidePhys} with AdamW optimizer~\citep{loshchilov2017decoupled} for 15,000 training steps.
The learning rate and batch size are set to $1 \times 10^{-4}$ and 32 respectively.
The proposed physics-aware video generation are trained on \textit{CollidePhys} and \textit{CompoPhys} with the learning rate of $5 \times 10^{-6}$.
All experiments are conducted on a single NVIDIA RTX A6000 GPU.
\textbf{Evaluation Configurations.} 
We measure physical consistency with the Physical Invariance Score (PIS) of NewtonGen~\citep{yuan2025newtongen}, the relative stability of a motion invariant, normalized to $[0,1]$. Visual fidelity uses FID, FVD, PSNR, SSIM, and EPE~\citep{huang2023vbench,unterthiner2018fvd,huang2024vbenchpp,liu2024evalcrafter}, covering spatial quality, temporal quality, and trajectory error.

\subsection{Comparison with State-of-the-Art Methods}
\textbf{Comparison Configurations.}
General T2V baselines are Sora~\citep{openai2024sora}, Veo3~\citep{wiedemer2025videomodelszeroshotlearners}, CogVideoX-5B~\citep{yang2025cogvideox}, Wan2.2~\citep{wan2025wanopenadvancedlargescale}, and vanilla Wan-Move~\citep{chu2025wanmove}. Physics-constrained baselines are retrained NewtonGen~\citep{yuan2025newtongen} and PhysT2V~\citep{xue2025phyt2vllmguidediterativeselfrefinement}. 
%
%

%
\textbf{Quantitative Results.}
We evaluate physical consistency following the standard protocol of NewtonGen~\citep{yuan2025newtongen} on 9 motion categories, covering 7 composite motions and 2 elastic collisions, and report per-parameter PIS in Table~\ref{tab:pis_comparison} (best and second-best highlighted).
CompAdapt obtains the highest PIS on every reported parameter. General T2V models score lower because they rely on visual correlations rather than explicit dynamics. Retrained NewtonGen ranks second on composite motions, and the remaining gap reflects explicit composite motion modeling; on elastic collisions it cannot produce valid videos, whereas CompAdapt stays close to the simulation reference. 

\begin{figure*}[t]
  \centering
  \includegraphics[width=\textwidth]{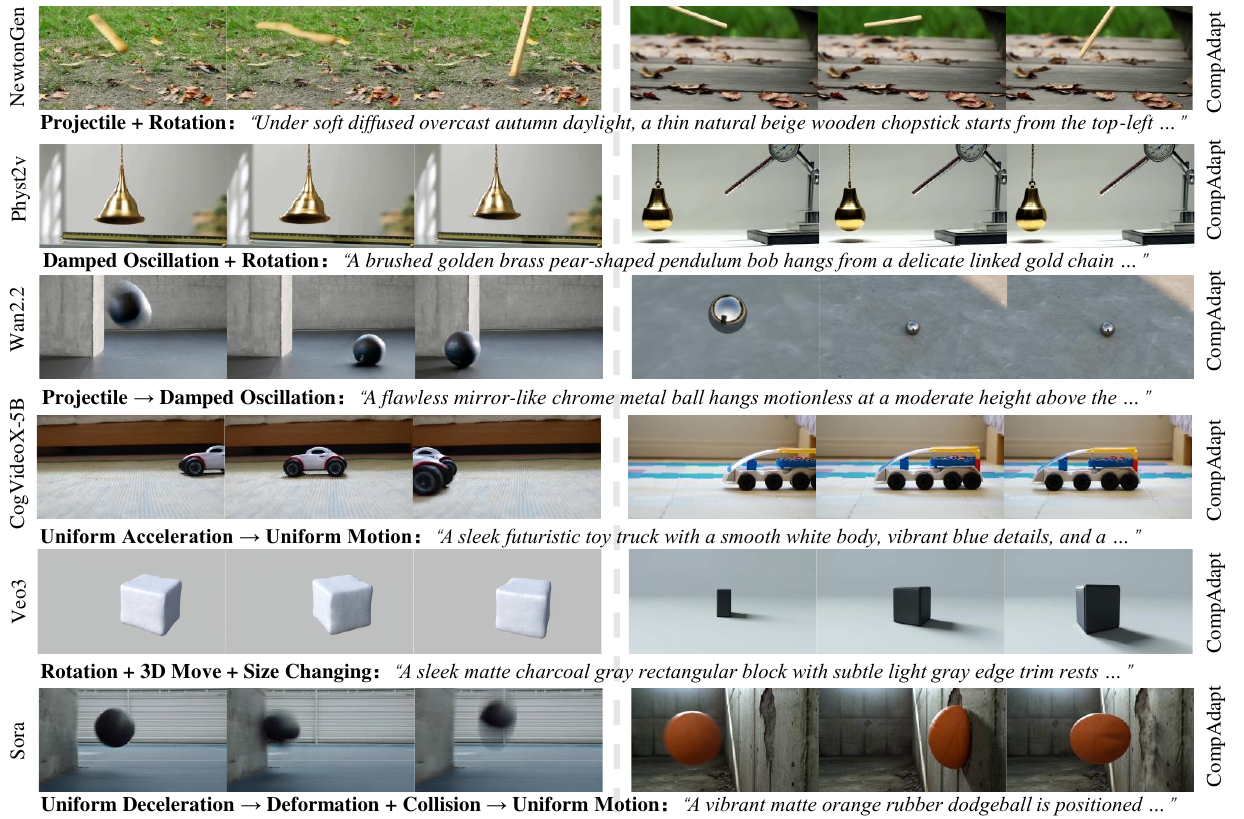}
  \caption{Qualitative comparison results on some representative physical motion scenarios. 
  Our CompAdapt generates videos with accurate physical motion and consistent appearance.
  }
  \vspace{-4mm}
  \label{fig:qualitative_comparison}
\end{figure*}

\textbf{Qualitative Results.}
In \cref{fig:qualitative_comparison}, general T2V models produce implausible trajectories, and physics-constrained baselines blur textures. CompAdapt keeps both the physical trajectory and the object appearance.

\begin{table*}[t!]
\centering
\caption{Per-parameter quantitative comparison of physical consistency (PIS). Reference is calculated on simulated ground-truth videos. * denotes retrained version.}
\label{tab:pis_comparison}
\resizebox{\textwidth}{!}{
\begin{tabular}{llcccccccc}
\toprule
\multirow{2}{*}{Motion Type} & \multirow{2}{*}{PIS$\uparrow$} & \multicolumn{8}{c}{Methods} \\
\cmidrule(lr){3-10}
& & Reference & Sora & Veo3 & CogVideoX-5B & Wan2.2 & PhysT2V & NewtonGen* & Ours \\
\midrule
\multicolumn{10}{c}{\textbf{\textit{Composite Motions}}} \\
\midrule
\multirow{3}{*}{\begin{tabular}[c]{@{}c@{}} Projectile + \\ Synchronous Rotation \end{tabular}} & $v_x$ & 0.9988 & 0.6548 & 0.7621 & 0.5392 & 0.6395 & 0.3474 & \secondbest{0.8914} & \best{0.9803} \\
& $a_y$ & 0.9487 & 0.5723 & 0.6187 & 0.4230 & 0.5571 & 0.3567 & \secondbest{0.7802} & \best{0.8189} \\
& $\omega$ & 0.9829 & 0.4267 & 0.5285 & 0.3380 & 0.3425 & 0.4119 & \secondbest{0.8905} & \best{0.9289} \\
\midrule
\multirow{2}{*}{\begin{tabular}[c]{@{}c@{}} Uniform Acceleration \\+ Synchronous Scaling\end{tabular}} & $a_x$ & 0.8489 & 0.3437 & 0.5033 & 0.4010 & 0.3077 & 0.5012 & \secondbest{0.6012} & \best{0.6568} \\
& $\Delta_r$ & 0.8501 & 0.2840 & 0.4167 & 0.4774 & 0.1972 & 0.3410 & \secondbest{0.6105} & \best{0.6362} \\
\midrule
\multirow{2}{*}{\begin{tabular}[c]{@{}c@{}} Circular Motion $\rightarrow$\\ Post-collision Linear\end{tabular}}& $\omega$ & 0.9933 & 0.6391 & 0.7726 & 0.5392 & 0.4677 & 0.6012 & \secondbest{0.8807} & \best{0.9788} \\
& $v$ & 0.9972 & 0.5349 & 0.6395 & 0.4774 & 0.5285 & 0.4310 & \secondbest{0.9784} & \best{0.9830} \\
\midrule
\multirow{2}{*}{\begin{tabular}[c]{@{}c@{}} Oblique Parabola $\rightarrow$ \\ Damped Oscillation\end{tabular}} & $v_x$ & 0.9988 & 0.6370 & 0.7747 & 0.5392 & 0.6187 & 0.4010 & \secondbest{0.8914} & \best{0.9803} \\
& $a_y$ & 0.9402 & 0.2841 & 0.3494 & 0.3083 & 0.3516 & 0.4418 & \secondbest{0.5107} & \best{0.5240} \\
\midrule
\multirow{4}{*}{\begin{tabular}[c]{@{}c@{}} Approaching + Rotation \\ Post-collision Rebound\end{tabular}} & $\Delta_l$ & 0.7388 & 0.2911 & 0.4583 & 0.3026 & 0.5013 & 0.5932 & \secondbest{0.6214} & \best{0.6472} \\
& $\omega$ & 0.9836 & 0.4267 & 0.5285 & 0.6596 & 0.3425 & 0.7842 & \secondbest{0.8593} & \best{0.8827} \\
& $v_y$ & 0.9986 & 0.6510 & 0.8384 & 0.6690 & 0.8481 & 0.8913 & \secondbest{0.9205} & \best{0.9371} \\
& $a_y$ & 0.9487 & 0.3567 & 0.5571 & 0.4230 & 0.5723 & 0.7662 & \secondbest{0.7802} & \best{0.8189} \\
\midrule
\multirow{3}{*}{\begin{tabular}[c]{@{}c@{}} Uniform Acceleration + \\ Scaling $\rightarrow$ Uniform Motion\end{tabular}} & $a_x$ & 0.8489 & 0.3437 & 0.5033 & 0.4010 & 0.3077 & 0.5012 & \secondbest{0.6012} & \best{0.6568} \\
& $\Delta_r$ & 0.8501 & 0.2840 & 0.4167 & 0.4774 & 0.1972 & 0.3410 & \secondbest{0.6105} & \best{0.6362} \\
& $v$ & 0.9972 & 0.5349 & 0.6395 & 0.4774 & 0.5285 & 0.4310 & \secondbest{0.9784} & \best{0.9830} \\
\midrule
\multirow{3}{*}{\begin{tabular}[c]{@{}c@{}}Damped Oscillation + Rotation \\$\rightarrow$ Uniform Translation\end{tabular}} & $a_y$ & 0.9402 & 0.2841 & 0.3494 & 0.3083 & 0.3516 & 0.4418 & \secondbest{0.5107} & \best{0.5240} \\
& $\omega$ & 0.9836 & 0.4267 & 0.5285 & 0.6596 & 0.3425 & 0.7842 & \secondbest{0.8601} & \best{0.8831} \\
& $v$ & 0.9972 & 0.5349 & 0.6395 & 0.4774 & 0.5285 & 0.4310 & \secondbest{0.9784} & \best{0.9830} \\
\midrule
\multicolumn{10}{c}{\textbf{\textit{Elastic Collision Motions}}} \\
\midrule
\multirow{4}{*}{\begin{tabular}[c]{@{}c@{}}2-object Direct \\Elastic Collision\end{tabular}} & $v_{x1}$ & 0.9891 & 0.4812 & 0.4675 & 0.4563 & 0.4704 & \secondbest{0.5217} & - & \best{0.9315} \\
& $v_{y1}$ & 0.9876 & 0.4784 & 0.4591 & 0.4485 & 0.4672 & \secondbest{0.5182} & - & \best{0.9287} \\
& $v_{x2}$ & 0.9884 & 0.4905 & 0.4703 & 0.4617 & 0.4736 & \secondbest{0.5279} & - & \best{0.9294} \\
& $v_{y2}$ & 0.9869 & 0.4847 & 0.4656 & 0.4552 & 0.4691 & \secondbest{0.5225} & - & \best{0.9276} \\
\midrule
\multirow{4}{*}{\begin{tabular}[c]{@{}c@{}}2-object Oblique \\ Elastic Collision\end{tabular}} & $v_{x1}$ & 0.9857 & 0.4773 & 0.4602 & 0.4531 & 0.4665 & \secondbest{0.5154} & - & \best{0.9182} \\
& $v_{y1}$ & 0.9842 & 0.4715 & 0.4564 & 0.4493 & 0.4627 & \secondbest{0.5108} & - & \best{0.9157} \\
& $v_{x2}$ & 0.9851 & 0.4802 & 0.4637 & 0.4575 & 0.4689 & \secondbest{0.5191} & - & \best{0.9169} \\
& $v_{y2}$ & 0.9838 & 0.4756 & 0.4598 & 0.4514 & 0.4643 & \secondbest{0.5146} & - & \best{0.9148} \\
\bottomrule
\end{tabular}
}
\vspace{-2mm}
\end{table*}

\begin{table*}[t!]
\centering

\begin{minipage}[t]{0.48\textwidth}
\centering
\captionof{table}{Effect of composite motion modeling.}
\label{tab:composite_ablation}
\resizebox{\linewidth}{!}{
\begin{tabular}{lccccc}
\toprule
Methods & PIS $\uparrow$ & FID $\downarrow$ & FVD $\downarrow$ & PSNR $\uparrow$ & SSIM $\uparrow$ \\
\midrule
w/o MTC & 0.615 & 18.7 & 102.3 & 15.1 & 0.55 \\
w/o Temporal & 0.589 & 19.2 & 105.7 & 14.8 & 0.54 \\
w/o Dual LLM & 0.724 & 16.8 & 96.2 & 16.3 & 0.58 \\
Ours & 0.862 & 15.0 & 91.4 & 17.2 & 0.61 \\
\bottomrule
\end{tabular}
}
\end{minipage}
\hfill
\begin{minipage}[t]{0.48\textwidth}
\centering
\captionof{table}{Comparison of video generators.}
\label{tab:wanmove_ablation}
\resizebox{\linewidth}{!}{
\begin{tabular}{lccccc}
\toprule
Methods & FID $\downarrow$ & FVD $\downarrow$ & PSNR $\uparrow$ & SSIM $\uparrow$ & EPE $\downarrow$ \\
\midrule
Go-with-the-Flow & 17.5 & 96.7 & 14.9 & 0.56 & 3.2 \\
Wan-Move & 16.3 & 93.5 & 16.5 & 0.58 & 2.9 \\
Ours & 15.0 & 91.4 & 17.2 & 0.61 & 2.9 \\
\bottomrule
\end{tabular}
}
\end{minipage}
\vspace{-4mm}
\end{table*}

\subsection{Ablation Studies}
\label{sec:ablation}

%

\textbf{Effect of Composite Motion Modeling.}
The \textit{w/o MTC} variant runs every component sequentially, the \textit{w/o Temporal} variant runs every component synchronously, and the \textit{w/o Dual LLM} variant uses one LLM for both decomposition and parameter prediction.
%
%
Table~\ref{tab:composite_ablation} shows that each removal lowers average PIS, so parallel composition, temporal chaining, and the decoupled parser contribute complementary information.


\textbf{Effect of Physics-Aware Video Generation.}
Table~\ref{tab:wanmove_ablation} shows that the spatial warp and adaptive blend improve visual quality over both Go-with-the-Flow~\citep{burgert2025go} and untuned Wan-Move~\citep{chu2025wanmove}. Corresponding frames appear in \cref{fig:wan_ablation}.
\textbf{Effect of One-shot Adaptation.}
For each \textit{OODPhys} scenario, we fine-tune the prior-matched module on a single reference (Eq.~\eqref{eq:adapt_loss}) and evaluate on the remaining trajectories. Table~\ref{tab:one_shot_tiers} shows high consistency in the near- and mid-domain.
Far-domain scenarios define an ambitious target for extending the module library, and per-scenario results with detailed regime analyses are provided in \cref{sec:app-oneshot,sec:app-oodphys}.
\begin{table}[!ht]
\centering
\caption{One-shot adaptation on \textit{OODPhys} (PIS $\uparrow$; mean and std over 5 reference trajectories).}
\label{tab:one_shot_tiers}
\renewcommand{\arraystretch}{0.9}
\scalebox{0.9}{\begin{tabular}{lccc}
\toprule
Difficulty Tier & \# Scenarios & Avg. Var-PIS & Avg. Std \\
\midrule
Near-domain (parameter shift) & 22 & 0.888 & 0.013 \\
Mid-domain (partial structure shift) & 5 & 0.854 & 0.015 \\
Far-domain (structural OOD) & 5 & 0.515 & 0.042 \\
Adversarial (capacity limit) & 2 & 0.730 & 0.009 \\
\midrule
Lunar free-fall (NASA footage) & 1 & 0.938 & 0.012 \\
Curling low-friction sliding (sports broadcast) & 1 & 0.921 & 0.015 \\
Damped pendulum (physics laboratory) & 1 & 0.907 & 0.013 \\
\bottomrule
\end{tabular}}
\vspace{-4mm}
\end{table}

\section{Conclusion}
We presented CompAdapt, which brings structured physical semantics and explicit dynamics into text-to-video generation for composite motion, collisions, and unseen physical laws.
Experiments show stronger physical consistency than general T2V and physics-constrained methods while preserving visual fidelity.
The planar, multi-object focus admits analytical guarantees.

\subsection*{AI use statement}
In this work, we did not use generative AI tools for any of the tasks that require disclosure under the ICLR 2027 AI Policy for Authors. The composite motion formulation, the dynamics and collision modules, the one-shot adaptation objective, the benchmark construction, and the interpretation of the experimental results are our own.

Among the tasks for which disclosure is recommended, we used generative AI tools to aid and polish the writing of this manuscript, specifically to edit the text for readability and to reformat the manuscript and its reference list for the ICLR 2027 template. We have reviewed all AI-assisted work: each edit was checked against our intended technical meaning, and no experimental result, claim, or reference was generated or substantively altered by an AI tool. We take responsibility for the final content of this work, including text, claims, and artifacts produced with the aid of generative AI.

\subsection*{Reproducibility statement}
The model formulation and training objectives are described in Section~\ref{sec:method}, while the datasets, evaluation protocols, baselines, and implementation settings are documented in Section~\ref{sec:experiments} and the appendix. The appendix additionally reports physical-state encoding details, parser evaluation, a theoretical justification of composite motion modeling, per-scenario one-shot adaptation results, benchmark construction, implementation details, and extended ablations to support reproduction and verification of the reported findings.

\clearpage

\bibliographystyle{iclr2027_conference}
\bibliography{componewton_references}

\clearpage
\appendix

\section*{Appendix}

The content of the supplementary material involves:
\begin{itemize}
\item[--] Physical parameter encoding in the 10D state in Sec.~\textbf{A}.
\item[--] More evaluation of the text-to-physical parser in Sec.~\textbf{B}.
\item[--] A theoretical justification of composite motion modeling in Sec.~\textbf{C}.
\item[--] Extended one-shot adaptation evaluation in Sec.~\textbf{D}.
\item[--] More details of the \textit{OODPhys} benchmark in Sec.~\textbf{E}.
\item[--] The Physical Invariance Score and implementation details in Sec.~\textbf{F}.
\item[--] Generalization to open-ended prompts in Sec.~\textbf{G}.
\item[--] Additional ablations, collision-time detection, and the video-generator comparison in Sec.~\textbf{H}.
\item[--] Full text prompts of the qualitative comparison in Sec.~\textbf{I}.
\end{itemize}

\section{Physical Parameter Encoding in the 10D State}
\label{sec:app-state-probe}
The 10D state explicitly represents per-frame kinematics, while material and environmental properties (friction, damping, viscosity, stiffness, \etc) are implicitly encoded in the learnable parameters of the NND modules.
To verify that such properties are genuinely learned rather than absorbed by trajectory fitting, we probe the parameter encoding at two levels.
At the \textit{parameter level}, we train NND modules from scratch under systematically varying values of 8 physical quantities and compute the Pearson correlation between the learned ODE parameters and the ground-truth physical values.
At the \textit{trajectory level}, for each quantity we train a module on 40 trajectories under a fixed parameter value and evaluate it on (i) 10 unseen trajectories with the same parameter but different initial conditions and (ii) 10 trajectories under a distinctly different parameter value, reporting the PIS gap to the ground-truth trajectories and the MSE ratio between the incorrect- and correct-parameter settings.

\begin{table}[ht]
\centering
\caption{Correlation between learned ODE parameters and ground-truth physical values across 8 probed quantities (each probed at 7 values, 50 trajectories per value).}
\label{tab:probe_corr}
\scalebox{0.9}{\begin{tabular}{llcc}
\toprule
Physical Quantity & Encoding Mechanism & Pearson $|r|$ & Encoding \\
\midrule
Friction $\mu_f$ & 6 linear ODE parameters jointly & 0.999 & Strong \\
Oscillation frequency $f$ & second-order stiffness term & 1.000 & Strong \\
Size expansion rate & size ODE constant term $\beta_s$ & 0.989 & Strong \\
Area conservation under compression & emergent $\alpha_s + \alpha_l \approx 0$ & 0.980 & Strong \\
Gravitational acceleration $g$ & constant acceleration term $c_y$ & 0.914 & Strong \\
Damping coefficient & dedicated damping parameter & 0.817 & Moderate \\
Fluid viscosity $\eta$ & velocity damping term $b_y$ & 0.754 & Moderate \\
Spring stiffness $k$ & position- and velocity-dependent terms & 0.630 & Moderate \\
\bottomrule
\end{tabular}}
\vspace{-2mm}
\end{table}

\begin{table}[ht]
\centering
\caption{Trajectory-level generalization of the implicitly encoded parameters. PIS Gap is the difference between the PIS of the correct-parameter module and the ground-truth trajectories on unseen initial conditions; MSE Ratio compares the incorrect-parameter setting against the correct-parameter setting.}
\label{tab:probe_traj}
\scalebox{0.9}{\begin{tabular}{lccc}
\toprule
Physical Quantity & PIS Gap $\downarrow$ & MSE Ratio $\uparrow$ & Tier \\
\midrule
Gravitational acceleration $g$ & 0.003 & 174$\times$ & Precise encoding \\
Size expansion rate & 0.045 & 82$\times$ & Precise encoding \\
Area conservation under compression & 0.000 & 54$\times$ & Precise encoding \\
Spring stiffness $k$ & 0.022 & 51$\times$ & Solid encoding \\
Oscillation frequency $f$ & 0.002 & 38$\times$ & Solid encoding \\
Friction $\mu_f$ & 0.126 & 6$\times$ & Solid encoding \\
Damping coefficient & 0.001 & 3$\times$ & Weak specificity \\
Fluid viscosity $\eta$ & 0.128 & 1.6$\times$ & Weak specificity \\
\bottomrule
\end{tabular}}
\vspace{-2mm}
\end{table}

Three interpretable encoding patterns emerge from the probe.
First, \textit{direct mapping}, where a one-to-one correspondence exists between a physical quantity and an ODE parameter (\eg, the expansion rate maps to $\beta_s$, and the oscillation frequency maps to the stiffness term).
Second, \textit{multi-parameter joint encoding}, where a single property is distributed across several ODE components (\eg, friction is jointly represented by 6 linear parameters, consistent with friction affecting resultant acceleration).
Third, \textit{automatic constraint discovery}, where the NND satisfies conservation laws without explicit supervision (\eg, area conservation under compression emerges as $\alpha_s + \alpha_l \approx 0$, and the residual MLP self-zeros for linear motions).
\textbf{Scope and extensions.} As a tractable and well-defined instantiation, the framework is developed on planar dynamics, where the state blocks admit exact analytical treatment, and the collision module realizes two-body contacts through a closed-form state map. This choice deliberately factors out contact-surface mechanics, three-dimensional forces, and articulated bodies, which the compositional state-block design is well positioned to incorporate by adding coordinates---a natural direction for extension. The dominant numerical limits at high parameter values are set by ODE solver precision and frame sampling rather than model capacity, and are likewise relieved by finer integration.

\section{Text-to-Physical Parser Evaluation}
\label{sec:app-parser}
We evaluate the dual-LLM parser in isolation on a held-out test set of 2,000 annotated samples from \textit{Text2Phys}, covering all 12 motion categories and three composite structures (parallel, sequential, and hybrid), with an average of 2.3 motion components per sample.
We compare against zero-shot Qwen2.5-7B-Instruct~\citep{yang2024qwen25} and a single-LLM fine-tuned baseline that performs both decomposition and parameter prediction.

\textbf{Motion decomposition.}
As shown in Table~\ref{tab:parser_decomp}, the dedicated $\text{LLM}_{\text{seq}}$ branch achieves 99.1\% sequence-level accuracy and 99.4\% per-motion category accuracy, outperforming the single-LLM baseline by 20.7 percentage points on sequence-level accuracy.
For samples with three or more sequential stages, sequence-level accuracy remains 98.3\%.

\begin{table}[ht]
\centering
\caption{Motion decomposition accuracy on the \textit{Text2Phys} test set (2,000 samples).}
\label{tab:parser_decomp}
\scalebox{0.9}{\begin{tabular}{lcc}
\toprule
Method & Sequence-level Acc. $\uparrow$ & Per-motion Category Acc. $\uparrow$ \\
\midrule
Zero-shot Qwen2.5-7B-Instruct & 54.7\% & 70.2\% \\
Single-LLM fine-tuned baseline & 78.4\% & 86.9\% \\
Ours (dual-LLM parser) & \textbf{99.1\%} & \textbf{99.4\%} \\
\bottomrule
\end{tabular}}
\vspace{-2mm}
\end{table}

\textbf{Physical parameter prediction.}
The $\text{LLM}_{\text{param}}$ branch adopts a two-stage design in which the LLM performs semantic-level recognition and a deterministic post-processing program generates the final numerical values.
The LLM distinguishes three input scenarios: explicit numerical values (extracted verbatim), qualitative adjectives (\eg, ``high speed'', ``heavy''; mapped to \textit{small}/\textit{medium}/\textit{large} labels and sampled from the corresponding physical intervals), and absent descriptions (outputting a \textit{none} token and sampling from a globally reasonable range).
Correctness is defined as $\leq 1\%$ relative error for explicit values, exact label match for adjectives, and correct \textit{none} detection for absent descriptions.
As shown in Table~\ref{tab:parser_param}, the parser exceeds 99\% accuracy in all three scenarios, providing a reliable foundation for downstream dynamics modeling.
Qualitative examples: for ``A small steel ball falls freely from a height of 2 meters with an initial horizontal velocity of 3 m/s'', the parser outputs \texttt{y=2.0, vx=3.0, mass=small}; for ``A heavy wooden block slides quickly down a rough slope'', it outputs \texttt{mass=large, initial\_velocity=large}; for ``A ball performs projectile motion and then bounces off the wall'', it correctly outputs \texttt{mass=none, initial\_velocity=none}.

\begin{table}[ht]
\centering
\caption{Physical parameter parsing accuracy on 2,000 mixed test samples.}
\label{tab:parser_param}
\scalebox{0.9}{\begin{tabular}{lcccc}
\toprule
Method & Overall $\uparrow$ & Explicit Num. $\uparrow$ & Adjective Class. $\uparrow$ & No-Description $\uparrow$ \\
\midrule
Zero-shot Qwen2.5-7B-Instruct & 51.8\% & 70.0\% & 46.5\% & 40.5\% \\
Single-LLM fine-tuned baseline & 81.7\% & 88.0\% & 81.5\% & 75.5\% \\
Ours & \textbf{99.3\%} & \textbf{99.8\%} & \textbf{99.2\%} & \textbf{99.0\%} \\
\bottomrule
\end{tabular}}
\vspace{-2mm}
\end{table}

\section{Theoretical Justification of Composite Motion Modeling}
\label{sec:app-comp-proof}
We justify the additive composition in Eq.~\eqref{eq:syn_unit} and the collision map in Eq.~\eqref{eq:collision} from first principles. Each component module is trained to reproduce the corresponding component dynamics, as quantified by the Physical Invariance Score. We then show that the composition introduces no additional error and preserves every conserved quantity. Write the component dynamics as a smooth vector field $\dot{Z}=F_j(Z)$ with flow $\Phi_t^{F_j}$, and recall that forces add, $\dot{Z}=f=\sum_j F_j$. The state partitions into the translation block $\{x,y,v_x,v_y\}$, rotation block $\{\theta,\omega\}$, shape block $\{s,l,a\}$, and the mass coordinate $\{\mu\}$. Let $P_j$ be the diagonal projector onto the block acted on by component $j$.

\textbf{Exactness under block decoupling.}
Call $F_j$ \emph{block-supported} if it acts only on its own block and depends only on coordinates within it, \ie, $P_{j^\perp}F_j\equiv 0$, $\partial F_j/\partial z_k=0\ (k\notin\mathcal{B}_j)$. If the blocks are pairwise disjoint and every $F_j$ is block-supported, then
\begin{equation}
\Phi_t^{f}(Z_{\mathrm{init}})
=
Z_{\mathrm{init}}+\sum_j \mathbf{r}\odot\bigl(\Phi_t^{F_j}(Z_{\mathrm{init}})-Z_{\mathrm{init}}\bigr).
\label{eq:app-exact}
\end{equation}
On block $\mathcal{B}_m$, all $F_j$ with $j\neq m$ vanish identically, so their flows leave the block at its initial value. The right-hand side of Eq.~\eqref{eq:app-exact} restricted to $\mathcal{B}_m$ therefore equals $\Phi_t^{F_m}|_{\mathcal{B}_m}$. Since $F_m$ depends only on $\mathcal{B}_m$, this block evolves autonomously and $f|_{\mathcal{B}_m}=F_m|_{\mathcal{B}_m}$, and by uniqueness of ODE solutions it equals $\Phi_t^f|_{\mathcal{B}_m}$. The mask keeps $\mu$ fixed.

\textbf{Linear superposition on shared coordinates.}
When components share coordinates and the dynamics is affine linear, $\dot{x}=Ax+b$ with $b=\sum_j b_j$, the exact composite is the free evolution counted once plus the sum of the particular responses,
\begin{equation}
x(t)=e^{At}x_0+\sum_j x_{p,j}(t),\qquad \dot{x}_{p,j}=Ax_{p,j}+b_j,\quad x_{p,j}(0)=0,
\label{eq:app-linear}
\end{equation}
whereas the naive displacement sum repeats the free drift once per component (\eg two concurrent constant accelerations with nonzero initial velocity duplicate the $v_0t$ term). The particular-solution form, which a force-type module can be trained to output, is exact for all linear systems, including linear (Stokes) drag, damped and driven oscillators, and gyroscopic (Coriolis-type) coupling. The same result extends to the linearized dynamics of every system near a stable equilibrium: the quadratic expansion and the normal-mode transform $Ka=\omega^2Ma$ diagonalize those dynamics into independent harmonic oscillators, covering small-angle pendulums, coupled springs, and elastic deformation modes, while the higher-order nonlinear residual is controlled by Eq.~\eqref{eq:app-gronwall}.

\textbf{Preservation of invariants.}
If a quantity $I$ is conserved by every component, $\nabla I\cdot F_j=0$, it is conserved by the composite,
\begin{equation}
\frac{\mathrm{d}}{\mathrm{d}t}I=\nabla I\cdot\sum_j F_j=0;
\label{eq:app-invariant}
\end{equation}
block-supported fields likewise preserve each block's own invariant (momentum of the free block, angular momentum of the torque-free block, oscillator energy, and the area $a=\kappa sl$ with $\dot{a}/a=\dot{s}/s+\dot{l}/l=0$ under compression). Consequently the composition adds zero marginal PIS degradation and preserves every accurate module exactly.

\textbf{Collision map and the necessity of mass.}
Let $\mathbf{n}$ be the contact normal directed from $A$ to $B$, and $\boldsymbol{\ell}_A,\boldsymbol{\ell}_B$ the vectors from the centers of mass to the contact point, with planar cross product $\boldsymbol{\ell}\times\mathbf{n}=\ell_x n_y-\ell_y n_x$. Under the impulse pair $-J\mathbf{n},+J\mathbf{n}$ (Newton's third law), the impulse--momentum theorem gives
\begin{equation}
\begin{aligned}
\mathbf{v}_A^+&=\mathbf{v}_A^--\frac{J}{\mu_A}\mathbf{n}, & \mathbf{v}_B^+&=\mathbf{v}_B^-+\frac{J}{\mu_B}\mathbf{n},\\
\omega_A^+&=\omega_A^--\frac{J}{I_A}(\boldsymbol{\ell}_A\times\mathbf{n}), & \omega_B^+&=\omega_B^-+\frac{J}{I_B}(\boldsymbol{\ell}_B\times\mathbf{n}).
\end{aligned}
\label{eq:app-impulse}
\end{equation}
The contact-point normal approach speed and the effective inverse mass are
\begin{equation}
u^-=\mathbf{n}\cdot\mathbf{v}_A^-+\omega_A^-(\boldsymbol{\ell}_A\times\mathbf{n})-\mathbf{n}\cdot\mathbf{v}_B^--\omega_B^-(\boldsymbol{\ell}_B\times\mathbf{n}),\quad
W=\frac{1}{\mu_A}+\frac{1}{\mu_B}+\frac{(\boldsymbol{\ell}_A\times\mathbf{n})^2}{I_A}+\frac{(\boldsymbol{\ell}_B\times\mathbf{n})^2}{I_B},
\label{eq:app-W}
\end{equation}
with $u^+=u^--JW$. The restitution law $u^+=-e\,u^-$ uniquely determines $J=(1+e)u^-/W$, since $W>0$. The equal-and-opposite impulses act at one point, so total linear and angular momentum are conserved. Summing the kinetic-energy change over both bodies yields
\begin{equation}
T^+-T^-=-\frac{(1-e^2)(u^-)^2}{2W}\le 0,
\label{eq:app-energy}
\end{equation}
which vanishes for the elastic case $e=1$. At fixed contact geometry the map is affine linear in the pre-impact velocities, so the learned module $\mathcal{D}_{\mathrm{col}}$ can represent it exactly. Because $J$ depends on the masses through $W$, the mass coordinate is indispensable---identical pre-impact kinematics with different masses yield different post-impact states. The fixed-wall case follows at $\mu_B\to\infty$.

\textbf{Error bound under coupling.}
If the fields are $L$-Lipschitz on the state region traversed and the omitted coupling together with the module error satisfies $\|f-\sum_j F_j\|\le\eta$ throughout the interval, the Gr\"onwall inequality bounds the composition error by
\begin{equation}
\|\hat{Z}(t)-Z(t)\|\le \frac{\eta}{L}\bigl(e^{Lt}-1\bigr),
\label{eq:app-gronwall}
\end{equation}
which vanishes as the modules become accurate, the coupling is modeled, and the segment length shrinks. Temporal chaining accumulates this bound segment by segment. Additive composition is therefore exact for the quasi-independent families above---rigid-body kinematics, together with linear dissipation and small oscillations when force modules output the particular responses of Eq.~\eqref{eq:app-linear}---and controlled by Eq.~\eqref{eq:app-gronwall} when components are coupled, as quantified for the nonholonomic rolling constraint $v=\omega r$ in \cref{sec:app-ablations}. A constraint-aware extension follows by introducing Lagrange multipliers, $M\ddot{q}=\mathbf{F}-A^{T}\lambda$ with $A\ddot{q}=b$.

\section{Extended One-Shot Adaptation Evaluation}
\label{sec:app-oneshot}

\textbf{Protocol.}
Each \textit{OODPhys} scenario contains 10 trajectories (84 frames, $\mathrm{d}t{=}0.01\,\text{s}$).
For one-shot adaptation, we extract the reference trajectory, perform dynamics-aware prior matching over the 12 pre-trained modules by trajectory MSE (Eq.~\eqref{eq:adapt_loss} initialization), fine-tune the matched module on the single reference with the regularized loss in Eq.~\eqref{eq:adapt_loss}, and evaluate on the remaining 9 trajectories.
This is repeated for 5 different reference selections to obtain Var-PIS and the standard deviation.
For temporal extrapolation, the module is trained on the first 28 frames (one third) of the reference and evaluated on the full 84-frame sequence.

\textbf{Prior matching behavior.}
The prior selected by trajectory MSE frequently differs from the semantically matching module: lunar-gravity projectiles are best matched by the 3D-motion module, damped pendulums by the deformation module, viscous settling by the 3D-motion module, and spring oscillators by the acceleration or damped-oscillation modules (ZS-PIS column of Table~\ref{tab:oodphys_all}).
This validates dynamics-aware selection over semantic matching: the matched prior alone already provides a strong initialization, and fine-tuning then adapts it to the target environment.

\textbf{Sample sensitivity.}
Across the 34 scenarios, 24 have a standard deviation below 0.02 and 16 below 0.01, showing that one-shot adaptation is largely insensitive to the reference sample in near-domain and mid-domain scenarios.
The four scenarios with standard deviations above 0.03 (outward and tight spirals, pure rotation, and the combined lunar-plus-low-friction-plus-size-change scenario) are exactly those without a close pre-trained prior or with multi-factor parameter shifts.

\textbf{Temporal extrapolation.}
Training on the first third of frames and evaluating on the full sequence yields a PIS drop below 0.02 in 12 of 34 scenarios, indicating that adaptation learns reusable dynamical patterns rather than memorizing the reference trajectory.
Moderate drops (0.10--0.16) appear in five scenarios: the two spring oscillators, the two deceleration scenarios, and super gravity, where the early segment carries less information about the full dynamics.
In five further scenarios (pure rolling $\times 2$, rotation $\times 2$, and the double pendulum), PIS$_{\mathrm{out}}$ exceeds PIS$_{\mathrm{in}}$, \ie, the module is conservative under limited early information and remains stable in the extrapolated window.

\textbf{Behavior under extreme shifts.}
At $g{=}980$ (100$\times$ gravity), $\omega{=}15$, or $a{=}25$, the matched prior lies far outside its pre-training range. Physical Invariance Score and absolute trajectory error then report complementary aspects: the structural invariant remains moderately stable (0.83 for extreme gravity), while scale-dependent absolute error grows---confirming that PIS diagnoses invariant structure rather than scale accuracy, and that extreme regimes call for extending the parameter range of the module library.

\textbf{Real-world validation.}
We validate the full adaptation pipeline on three public real-world physics videos: a NASA lunar free-fall experiment, a curling broadcast with low-friction sliding, and a damped-pendulum physics laboratory recording.
For each video, SAM2~\citep{ravi2024sam} segments and tracks the target object, the state-extraction pipeline (\texttt{physical\_encoder}) converts the track into a physical state sequence, and one-shot adaptation proceeds exactly as in simulation.
Var-PIS is computed with the extracted real trajectory as reference, under natural imaging noise and tracking errors.
As shown in Table~\ref{tab:one_shot_tiers} of the main text, the adapted modules reach Var-PIS above 0.90 in all three cases with standard deviations below 0.016, confirming that the adaptation remains effective beyond controlled simulation.

\section{Details of the OODPhys Benchmark}
\label{sec:app-oodphys}
\textbf{Scenario tiers and overview.}
The expanded \textit{OODPhys} benchmark contains 34 controlled scenarios spanning seven dynamical categories and 3 real-world videos.
The categories cover shifted gravity, friction, damping, and thrust, coupled rolling, and structurally novel mechanisms including spring oscillation, spiral motion, and chaotic double pendulums.
Scenarios are grouped into four tiers by the structural gap between the target dynamics and the 12 pre-trained dynamics modules, as summarized in Table~\ref{tab:one_shot_tiers} of the main text.
The per-scenario results are listed in Table~\ref{tab:oodphys_all}.

\begin{table}[ht]
\centering
\caption{Per-scenario results of one-shot adaptation on the 34 controlled \textit{OODPhys} scenarios. GT-PIS is the PIS of the ground-truth simulation, \ie, the attainable upper bound. ZS-PIS is the PIS of the best matched pre-trained module without fine-tuning (prior matching only). Var-PIS and Std are the mean and standard deviation of the test PIS over 5 reference-sample selections. PIS$_{\mathrm{in}}$ and PIS$_{\mathrm{out}}$ are the PIS on the seen first third and the extrapolated remaining window when training on the first third of frames only.}
\label{tab:oodphys_all}
\resizebox{\textwidth}{!}{
\begin{tabular}{llcccccc}
\toprule
Scenario & Modified physics & GT-PIS & ZS-PIS & Var-PIS & Std & PIS$_{\mathrm{in}}$ & PIS$_{\mathrm{out}}$ \\
\midrule
\multicolumn{8}{c}{\textit{Near-domain: parameter shift (22)}} \\
\midrule
Lunar gravity projectile & $g{=}1.63$ & 1.000 & 0.972 & 0.982 & 0.005 & 0.981 & 0.969 \\
High gravity projectile & $g{=}12$ & 1.000 & 0.971 & 0.961 & 0.018 & 0.990 & 0.982 \\
Super gravity projectile & $g{=}30$ & 1.000 & 0.974 & 0.891 & 0.009 & 0.962 & 0.815 \\
Low-friction slope & $\mu_f{=}0.01$ & 1.000 & 0.998 & 0.982 & 0.015 & 0.993 & 0.982 \\
Standard slope & $\mu_f{=}0.1$ & 1.000 & 0.998 & 0.978 & 0.011 & 0.997 & 0.992 \\
Lunar slope & $g{=}1.63$ & 1.000 & 0.998 & 0.913 & 0.027 & 0.991 & 0.977 \\
Low-damping pendulum & $d{=}0.05$ & 0.989 & 0.947 & 0.895 & 0.006 & 0.961 & 0.928 \\
Standard-damping pendulum & $d{=}0.8$ & 0.880 & 0.947 & 0.896 & 0.006 & 0.961 & 0.929 \\
Overdamped pendulum & $d{=}3.0$ & 0.833 & 0.947 & 0.901 & 0.007 & 0.961 & 0.930 \\
Negative-damping pendulum & $d{=}-0.5$ & 0.878 & 0.947 & 0.894 & 0.006 & 0.961 & 0.928 \\
Lunar pendulum & $g{=}1.63$ & 0.857 & 0.947 & 0.936 & 0.024 & 0.961 & 0.931 \\
Standard acceleration & $a{=}4$ & 1.000 & 0.997 & 0.942 & 0.005 & 0.987 & 0.964 \\
Extreme acceleration & $a{=}15$ & 1.000 & 0.997 & 0.916 & 0.001 & 0.989 & 0.968 \\
Standard deceleration & $a{=}8$ & 1.000 & 0.997 & 0.574 & 0.028 & 0.937 & 0.792 \\
Extreme deceleration & $a{=}25$ & 0.600 & 0.997 & 0.574 & 0.028 & 0.935 & 0.779 \\
Pure rotation & $\omega{=}1.5$ & 1.000 & 0.805 & 0.753 & 0.042 & 0.796 & 0.895 \\
Fast rotation & $\omega{=}15$ & 1.000 & 0.823 & 0.737 & 0.006 & 0.780 & 0.836 \\
Constant size expansion & $\beta_s$ shift & 1.000 & 0.999 & 0.969 & 0.001 & 0.995 & 0.989 \\
Oscillating size & periodic $\beta_s$ & 0.106 & 0.999 & 0.955 & 0.006 & 0.992 & 0.981 \\
Standard thrust & $a{=}3$ & 1.000 & 0.990 & 0.987 & 0.002 & 0.990 & 0.984 \\
Strong thrust & $a{=}8$ & 1.000 & 0.990 & 0.987 & 0.002 & 0.990 & 0.984 \\
Combined lunar $g$ + low friction + size change & multi-factor & 1.000 & 0.998 & 0.904 & 0.032 & 0.991 & 0.977 \\
\midrule
\multicolumn{8}{c}{\textit{Mid-domain: partial structure shift (5)}} \\
\midrule
Viscous settling & $\eta{=}1.0$ & 0.736 & 0.959 & 0.940 & 0.029 & 0.981 & 0.975 \\
Viscous settling, high viscosity & $\eta{=}5.0$ & 0.621 & 0.959 & 0.925 & 0.021 & 0.919 & 0.893 \\
Non-uniform gravity & $g(y){=}g_0(R/(R{+}y))^2$ & 1.000 & 0.930 & 0.825 & 0.017 & 0.971 & 0.940 \\
Pure rolling on slope & $\mu_r{=}0.01$ & 0.899 & 0.844 & 0.794 & 0.004 & 0.837 & 0.903 \\
Pure rolling, high friction & $\mu_r{=}0.1$ & 0.909 & 0.844 & 0.784 & 0.002 & 0.837 & 0.903 \\
\midrule
\multicolumn{8}{c}{\textit{Far-domain: structural OOD (5)}} \\
\midrule
Spring oscillator & $k{=}10$ & 0.798 & 0.507 & 0.483 & 0.011 & 0.705 & 0.584 \\
Stiff spring oscillator & $k{=}50$ & 0.395 & 0.488 & 0.515 & 0.018 & 0.728 & 0.600 \\
Outward spiral & $\dot{r}{=}0.5$ & 0.562 & 0.565 & 0.423 & 0.082 & 0.417 & 0.317 \\
Tight spiral & $\dot{r}{=}2.0$ & 0.569 & 0.700 & 0.459 & 0.083 & 0.369 & 0.365 \\
Large-amplitude circular motion & large radius & 0.776 & 0.761 & 0.693 & 0.017 & 0.746 & 0.724 \\
\midrule
\multicolumn{8}{c}{\textit{Adversarial: capacity limit (2)}} \\
\midrule
Chaotic double pendulum & 4-DOF coupling & 0.647 & 0.643 & 0.629 & 0.000 & 0.628 & 0.633 \\
100$\times$ extreme gravity & $g{=}980$ & 1.000 & 0.946 & 0.830 & 0.017 & 0.966 & 0.942 \\
\bottomrule
\end{tabular}
}
\vspace{-2mm}
\end{table}

Two scenarios deserve explanation.
For standard and extreme deceleration, GT-PIS is bounded at 1.000 and 0.600 because the monitored invariant (constant acceleration) is inherently less stable for strongly decelerating motion, and Var-PIS closely tracks GT-PIS, \ie, adaptation reaches the theoretically attainable consistency.
For oscillating size, the ground-truth trajectory itself scores 0.106 on the size-change invariant, so the adapted module (0.955) exceeds the ground-truth score on this metric; we retain the scenario to document metric behavior rather than adaptation quality.

\textbf{Dynamics of the new scenario families.}
The far- and mid-domain scenarios introduce dynamical mechanisms absent from the pre-trained module library:
spring-mass oscillation, $m\ddot{x} + kx = 0$, with solution $x(t) = A\cos(\omega t + \varphi)$ and $\omega = \sqrt{k/m}$, whose invariant is the total spring energy $\tfrac{1}{2}k(x - x_{\mathrm{eq}})^2 + \tfrac{1}{2}mv^2$;
viscous settling under the three-force balance $m\ddot{y} = (m - \rho_f V)g - 6\pi\eta r\dot{y}$, converging to the terminal velocity $v_t = 2r^2(\rho_o - \rho_f)g/(9\eta)$;
non-uniform gravity $g(y) = g_0\,(R/(R+y))^2$, which generalizes the constant-$g$ assumption;
pure rolling with the no-slip constraint $v = \omega r$, yielding $a = g(\sin\theta - \mu_r\cos\theta)/(1 + I/(mr^2))$ and $a = \tfrac{2}{3}g(\sin\theta - \mu_r\cos\theta)$ for a solid cylinder;
constrained spiral motion $r(t) = r_0 + \dot{r}t$ with constant angular velocity;
and the chaotic double pendulum, a 4-DOF nonlinear coupled system that motivates extending the state representation beyond the current ten dimensions.

\textbf{Generation protocol.}
Each scenario contains 10 trajectories of 84 frames simulated at $\mathrm{d}t = 0.01\,\text{s}$ by integrating the target dynamics with randomized initial conditions (initial positions, velocities, and object sizes).
The 12 pre-trained single-object modules (uniform motion, acceleration, deceleration, parabolic motion, parabolic motion with rotation, circular motion, rotation, damped oscillation, slope sliding, size changing, deformation, and 3D motion) form the candidate set for prior matching.

\section{Implementation Details}
\label{sec:app-impl}
\textbf{Physical Invariance Score (PIS).}
Following NewtonGen~\citep{yuan2025newtongen}, the Physical Invariance Score measures the relative stability of a motion-invariant physical quantity $q$ over a trajectory and is normalized to $[0,1]$, \ie,
\begin{equation}
\mathrm{PIS} = \frac{1}{1 + \mathrm{std}(q) / (|\mathrm{mean}(q)| + \varepsilon)} .
\end{equation}
Each scenario is evaluated only on the invariant associated with its own physical law (\eg, horizontal velocity and energy for projectiles, pendulum energy for pendulum-like motions, the $v = \omega r$ coupling for pure rolling), so PIS reflects how well the generated trajectory obeys the target law.

\textbf{Hyperparameters and reproducibility.}
Table~\ref{tab:hyperparams} summarizes the key hyperparameters of each trainable component.
All experiments are conducted on a single NVIDIA RTX A6000 GPU.
All data generation, adaptation, and evaluation pipelines are deterministic, and every experiment uses fixed random seeds.
To ensure reproducibility, we will release the code, the datasets (\textit{Text2Phys}, \textit{CollidePhys}, \textit{CompoPhys}, and the expanded \textit{OODPhys}), the evaluation scripts, the state-extraction pipeline, and the generated videos upon acceptance.

\begin{table}[ht]
\centering
\caption{Key hyperparameters of the trainable components.}
\label{tab:hyperparams}
\scalebox{0.9}{\begin{tabular}{llll}
\toprule
Component & Setting & Value \\
\midrule
\multirow{4}{*}{Dual-LLM parser} & Backbone & Qwen2.5-7B-Instruct + LoRA \\
& LoRA rank / learning rate / batch size & 8 / $2{\times}10^{-5}$ / 8 \\
& Epochs & 3 \\
& Decoding & greedy, $\leq$ 512 tokens \\
\midrule
\multirow{3}{*}{Collision module $\mathcal{D}_{\mathrm{col}}$} & Optimizer & AdamW \\
& Learning rate / schedule & $1{\times}10^{-4}$ / cosine annealing \\
& Training steps / batch size & 15,000 / 32 \\
\midrule
\multirow{3}{*}{Physics-aware video generator} & Learning rate & $5{\times}10^{-6}$ \\
& Batch size & 1 \\
& Training data & \textit{CollidePhys} + \textit{CompoPhys} \\
\midrule
\multirow{3}{*}{One-shot adaptation} & Optimizer & AdamW \\
& Learning rate / schedule & $5{\times}10^{-3}$ / cosine annealing \\
& Epochs / trajectory length & 100 / 84 frames ($\mathrm{d}t{=}0.01\,\text{s}$) \\
\bottomrule
\end{tabular}}
\vspace{-2mm}
\end{table}

\section{Generalization to Open-Ended Prompts}
\label{sec:app-openprompts}
To test generalization beyond the controlled benchmark vocabulary, we collect 50 open-ended prompts with diverse object categories, materials, lighting, and scene contexts, with zero overlap with the original test set, and generate 10 videos per prompt.
As shown in Table~\ref{tab:openprompts}, the average PIS is 0.859 versus 0.862 on the \textit{CompoPhys} test set, and FID/FVD remain essentially unchanged at 15.0/91.5 versus 15.0/91.4.
Physical guidance therefore preserves both physical consistency and visual quality across diverse visual contexts, consistent with the strictly decoupled design of physical-semantics extraction and visual rendering: the parser is trained to filter out appearance modifiers, and the video backbone renders all visual content.
On the same prompts, the parser's decomposition accuracy decreases by at most 0.2\% relative to the in-distribution test set.

\begin{table}[ht]
\centering
\caption{Physical consistency and visual quality on 50 open-ended prompts versus the \textit{CompoPhys} test set.}
\label{tab:openprompts}
\scalebox{0.9}{\begin{tabular}{lccc}
\toprule
Setting & PIS $\uparrow$ & FID $\downarrow$ & FVD $\downarrow$ \\
\midrule
\textit{CompoPhys} test set & 0.862 & 15.0 & 91.4 \\
Open-ended prompts (50) & 0.859 & 15.0 & 91.5 \\
\bottomrule
\end{tabular}}
\vspace{-2mm}
\end{table}

\section{Additional Ablations}
\label{sec:app-ablations}

\begin{figure}[ht]
  \centering
  \includegraphics[width=\columnwidth]{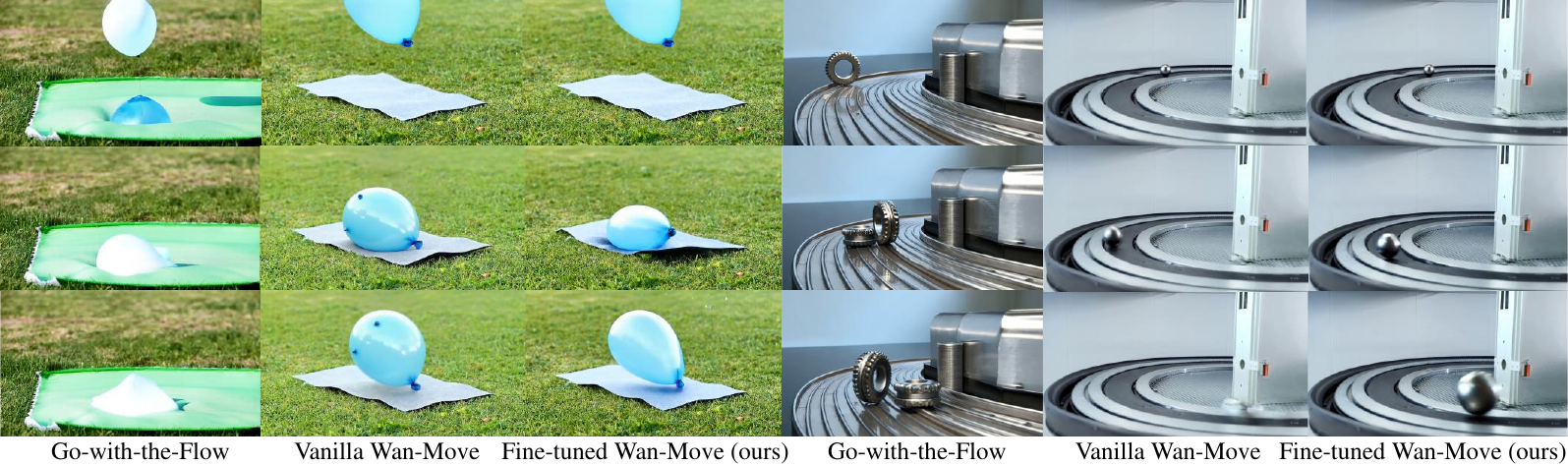}
  \caption{Qualitative comparison of different video generators. Our method maintains consistent visual appearance while ensuring accurate physical motion and sharper texture details.}
  \label{fig:wan_ablation}
\end{figure}

\textbf{Collision-time detection.}
Each object is modeled as a convex shape $\mathcal{K}$ parameterized by its centroid $\mathbf{c}_t = (x_t, y_t)$, orientation $\theta_t$, and dimensions $(s, l)$ extracted from $\mathbf{Z}_t$.
The minimum boundary distance used in Sec.~\ref{sec:comp_motion} is
\begin{equation}
d_t = \max_{\|\mathbf{d}\|=1}\; \bigl[\,\mathbf{d} \cdot (\mathbf{c}_t^{(B)} - \mathbf{c}_t^{(A)}) - h^{(A)}(\mathbf{d}) - h^{(B)}(-\mathbf{d})\,\bigr],
\label{eq:boundary_dist}
\end{equation}
where $h^{(n)}(\mathbf{d}) = \max_{\mathbf{x} \in \mathcal{K}^{(n)}} \,\mathbf{d} \cdot \mathbf{x}$ is the support function.
A collision is registered at $t_c = t$ when $d_t \leq \epsilon$, and $\mathcal{D}_{\mathrm{col}}$ is applied via Eq.~\eqref{eq:collision}.
When $d_t > \epsilon$, each object continues under the NND temporal chaining framework.

\textbf{Mass dimension and the MOT.}
The mass dimension $\mu$ and the collision-aware MOT are both prerequisite for physically correct collision modeling: without mass, the collision solver cannot enforce momentum conservation, and without the solver, colliding objects pass through each other with no interaction.
As shown in Table~\ref{tab:ablation_collision}, removing the mass dimension (9D state) reduces the average collision PIS from 0.923 to 0.689, and removing the MOT reduces it to 0.487, comparable to general T2V models.

\begin{table}[ht]
\centering
\caption{Ablations on multi-object elastic collisions (PIS $\uparrow$).}
\label{tab:ablation_collision}
\scalebox{0.9}{\begin{tabular}{lccc}
\toprule
Variant & Direct Collision & Oblique Collision & Avg \\
\midrule
w/o mass (9D state) & 0.694 & 0.683 & 0.689 \\
w/o MOT & 0.492 & 0.481 & 0.487 \\
Ours (10D state + MOT) & \textbf{0.929} & \textbf{0.916} & \textbf{0.923} \\
\bottomrule
\end{tabular}}
\vspace{-2mm}
\end{table}

\textbf{Scope of additive composition in the MTC.}
As shown in \cref{sec:app-comp-proof}, additive composition is exact when parallel components act on disjoint subsets of the state space, which covers the dominant composite motions in the benchmark---rotation paired with deceleration, scaling paired with translation. On shared coordinates, the free-evolution-corrected form of Eq.~\eqref{eq:syn_unit} is likewise exact for affine-linear dynamics, counting the force-free evolution once and adding each component's particular response. A residual appears only when the coupling is genuinely nonlinear or constrained. The scheme then supplies a well-defined approximation whose residual is governed by the Gr\"onwall bound (Eq.~\eqref{eq:app-gronwall}). Pure rolling illustrates this regime: the no-slip relation $v=\omega r$ links the translational and rotational blocks, so independent rollouts approximate rather than exactly realize the constraint. The residual is small, with adapted Var-PIS 0.784--0.794 against 0.978--0.982 for uncoupled slopes under comparable shifts, and 0.737 for rotation beyond the trained angular-velocity range ($\omega{=}15$). These measurements map the route forward: quasi-independent components already compose exactly, while constraint-aware extensions (Sec.~C) target coupled rolling, dense contact, and articulated chains.

\section{Full Text Prompts of the Qualitative Comparison}
\label{sec:app-qual-prompts}
Due to the limited space, each row of \cref{fig:qualitative_comparison} in the main paper only displays a truncated prefix of its text prompt.
For completeness and reproducibility, \cref{tab:qual_prompts} re-typesets all six prompts in full, listed in the same top-to-bottom order as the rows of \cref{fig:qualitative_comparison}.
Within each row, all compared methods are driven by exactly the same prompt, and no per-method prompt engineering is applied.

\begin{table}[!ht]
\centering
\caption{Full text prompts used in the qualitative comparison of \cref{fig:qualitative_comparison}. The rows follow the top-to-bottom order of the figure, and the truncated prefixes shown in the figure are reproduced here in full.}
\label{tab:qual_prompts}
\small
\begin{tabularx}{\textwidth}{@{}p{0.2\textwidth}X@{}}
\toprule
Motion Type & Full Text Prompt \\
\midrule
(1) Projectile + Synchronous Rotation & ``Under soft diffused overcast autumn daylight, a thin natural beige wooden chopstick starts from the top-left corner of the frame at a height of 0.9 meters above the deck. It is tossed diagonally downward at a 35-degree angle with a leisurely initial speed, traveling along a smooth parabolic path while rotating at a barely noticeable, glacial clockwise pace, barely turning a right angle throughout its entire descent. Below it lies a rustic weathered gray wooden deck scattered with curled, faded brown and amber fallen leaves, with a hazy backdrop of out-of-focus vibrant green bushes.'' \\
\addlinespace
(2) Damped Oscillation + Rotation & ``A brushed golden brass pear-shaped pendulum bob hangs from a delicate linked gold chain, initially held motionless at its leftmost swing position. It begins a gentle, rhythmic simple pendulum motion, swinging smoothly from side to side, and spins slowly and continuously clockwise as it moves. The scene is set on a pristine white laboratory surface, with a sleek silver precision gauge instrument featuring a circular dial and graduated scale resting to the right.'' \\
\addlinespace
(3) Projectile $\rightarrow$ Damped Oscillation & ``A flawless mirror-like chrome metal ball hangs motionless at a moderate height above the ground. It accelerates downward in a smooth vertical free fall, touches the surface, and then begins a soft, rhythmic damped simple harmonic oscillation.'' \\
\addlinespace
(4) Uniform Acceleration $\rightarrow$ Uniform Motion & ``A sleek futuristic toy truck with a smooth white body, vibrant blue details, and a sunny yellow roof rests motionless on the soft pastel-colored interlocking foam play mat. It begins accelerating steadily to the left, its black wheels spinning faster and faster with gentle momentum, before reaching a constant, smooth cruising speed and moving uniformly across the mat toward the left side of the frame.'' \\
\addlinespace
(5) Rotation + 3D Move + Size Changing & ``A sleek matte charcoal gray rectangular block with subtle light gray edge trim rests motionless at the center of a clean empty space. It glides steadily forward in fluid 3D motion toward the viewer, spins slowly and continuously clockwise around its central vertical axis, and grows larger evenly in all dimensions as it approaches. The scene features a seamless matte light gray floor and background, with soft directional lighting casting a gentle elongated shadow that shifts and scales in sync with the block's movement.'' \\
\addlinespace
(6) Uniform Deceleration $\rightarrow$ Deformation + Collision $\rightarrow$ Uniform Motion & ``A vibrant matte orange rubber dodgeball is positioned in the left half of an empty industrial concrete room, initially traveling steadily straight toward the right wall at a moderate constant speed. It strikes the textured gray concrete wall with a firm impact, squishes noticeably into an oblate shape at the moment of collision, then bounces cleanly off the wall and continues moving straight to the left at an identical steady speed, with motion blur emphasizing its continuous movement.'' \\
\bottomrule
\end{tabularx}
\vspace{-2mm}
\end{table}

\end{document}